\documentclass[10pt,twocolumn,letterpaper]{article}

\usepackage{cvpr}              
\usepackage[accsupp]{axessibility}
\usepackage{graphicx}
\usepackage{amsmath}
\usepackage{amssymb}
\usepackage{booktabs}
\usepackage[linesnumbered,ruled,vlined]{algorithm2e} 
\usepackage{setspace}
\usepackage[numbers,sort]{natbib}

\usepackage[pagebackref,breaklinks,colorlinks]{hyperref}

\usepackage[capitalize]{cleveref}
\crefname{section}{Sec.}{Secs.}
\Crefname{section}{Section}{Sections}
\Crefname{table}{Table}{Tables}
\crefname{table}{Tab.}{Tabs.}

\def\cvprPaperID{*****} 
\def\confName{CVPR}
\def\confYear{2022}

\begin{document}

\title{Ice hockey player identification via transformers and weakly supervised learning}

\author{Kanav Vats  \quad  \text{William McNally}  \quad  \text{Pascale Walters}$^{\dag}$ \quad  \\ \text{David A. Clausi} \quad  \text{John S. Zelek}\\
Systems Design Engineering, University of Waterloo \quad\quad $^{\dag}$Stathletes Inc \hspace{2cm}\\
{\tt\small \{k2vats, wmcnally, dclausi,jzelek\}@uwaterloo.ca} \quad\quad {\tt\small pascale.walters@stathletes.com}
}

\maketitle

\begin{abstract}
 Identifying players in video is a foundational step in computer vision-based sports analytics. Obtaining player identities is essential for analyzing the game and is used in downstream tasks such as game event recognition. Transformers are the existing standard in natural language processing (NLP) and are swiftly gaining traction in computer vision. Motivated by the increasing success of transformers in computer vision, we introduce a transformer network for recognizing players through their jersey numbers in broadcast National Hockey League (NHL) videos. The transformer takes temporal sequences of player frames (called player tracklets) as input and outputs the probabilities of jersey numbers present in the frames. The proposed network performs better than the previous benchmark on the same dataset. We implement a weakly-supervised training approach by generating approximate frame-level labels for jersey number presence and use the frame-level labels for faster training. We also utilize player shifts available in the NHL play-by-play data by reading the game time using optical character recognition (OCR) to get the players on the ice rink at a certain game time. Using player-shifts  improved the player identification accuracy   by $6\%$.

\end{abstract}

\section{Introduction}

Player identification is a problem of fundamental importance in vision-based sports analytics. Identifying players is a key component of player tracking systems \cite{vats2021player,jjl}  that are used by hockey coaches, analysts, and scouts to analyze the game.  \\

\begin{figure*}[t]
\begin{center}
\includegraphics[width = \linewidth]{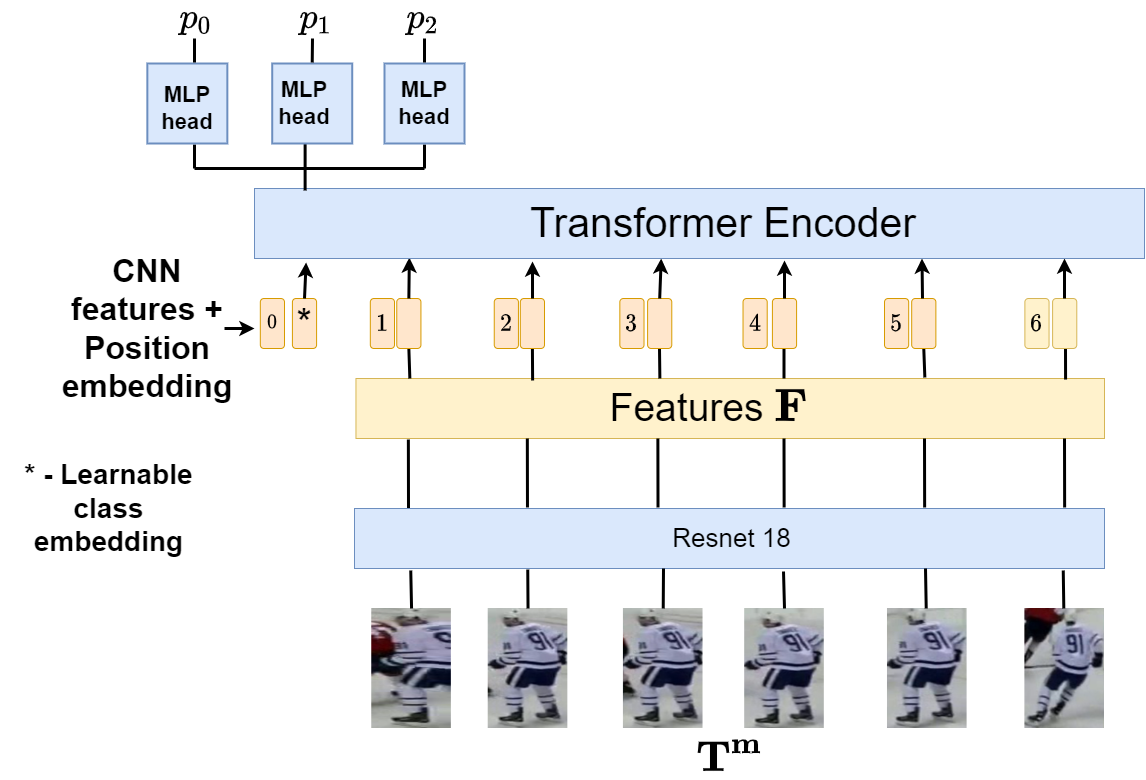}
\end{center}
  \caption{Network architecture for the proposed network. The input to the network is a temporal sequence of $m$ images $\mathbf{T^m}$. Each image in the tracklet is passed through a ResNet18 network to obtain 512 dimensional features $\mathbf{F}$. The features are prepended with the $[\texttt{class}]$ token and combined with learnable positional encoding. }
\label{fig:overall_arch}
\end{figure*}

Player identification through jersey numbers has been performed using static images \cite{gerke,li,liu,vatsmmsports}. However, inferring jersey number from static images does not take into account the valuable temporal information present in sports videos. To address the issue, Chan \textit{et al.} \cite{CHAN2021113891} and Vats \textit{et al.} \cite{vats2021player} infer jersey numbers from temporal player sequences called tracklets using an LSTM and temporal 1D CNN, respectively.  Inspired by the increasing success of transformers in computer vision tasks involving both  images \cite{detr, dosovitskiy2020vit,Li_2021_CVPR} and  videos \cite{Gavrilyuk2020ActorTransformersFG, vatn,Arnab_2021_ICCV}, in this paper, we introduce a transformer for recognizing jersey numbers from  player tracklets. The transformer takes as input CNN features of tracklet frames combined with a positional encoding and outputs the probabilities of jersey numbers present in the tracklet. We use the multi-task loss function proposed in Vats \textit{et al.} \cite{vatsmmsports} for training the network. The overall network is illustrated in Fig. \ref{fig:overall_arch}. The transformer network shows better performance compared to the previous benchmark on the same player identification dataset \cite{vats2021player}. \\

One detail common in Chan \textit{et al.} \cite{CHAN2021113891}  and Vats \textit{et al.} \cite{vats2021player} is that all images in a tracklet are annotated with the same label and a tracklet consists of hundreds of frames. As a result, when sampling a fixed number of frames for training, it is possible that the frames may not have a jersey number visible. This leads to inconsistent and slow training. In this paper, we perform weakly-supervised training by generating approximate frame-level labels for tracklet jersey numbers, which leads to faster training.\\

 For further improvement of player identification, we exploit the public NHL play-by-play data that contains information about which players are on the ice at any time of the game. Although the number of players on an NHL team roster is 23, there can be only between 3 and 5 players on the ice for each team at any point in the game (plus one goalie per team). We process this information using an optical character recognition (OCR) system that reads the game time and extracts the players on the ice using a player shift database. We multiply the final jersey number probability vector of a tracklet by a binary vector that encodes which players are present on the ice at a certain time. Using player shift information improves the overall accuracy by $6\%$.  The following items summarize the contributions of this paper:\\

\begin{enumerate}
    \item We introduce a weakly-supervised training strategy by obtaining approximate frame-wise jersey labels from a secondary network. The training strategy achieves faster convergence when compared to the naive strategy of not using approximate labels.
    \item We introduce a network composed of a transformer encoder for sports jersey number recognition that performs better than the previous benchmark on the dataset \cite{vats2021player}.
    \item We incorporate player shift times into the inference using OCR, allowing the network to focus on the players present at a particular moment in a game. Using player shifts improves player identification accuracy by a further $6\%$.
\end{enumerate}


\section{Background}

\subsection{Transformers in computer vision}

Following the success of the attention mechanism used in the NLP Transformer~\cite{vaswani2017attention}, many computer vision researchers have opted to incorporate elements of the Transformer into their architectures for image and video recognition. Many of the earlier approaches used CNN feature extractors with a Transformer-based network head. Girdhar \etal ~\cite{vatn} re-purposed the Transformer architecture for video understanding using a custom multi-head attention unit to process spatio-temporal features that were extracted using an I3D base. Their approach achieved state-of-the-art accuracy on the Atomic Visual Actions dataset~\cite{gu2018ava} using only RGB input. In a similar manner, Gavrilyuk \etal ~\cite{Gavrilyuk2020ActorTransformersFG} used a Transformer encoder to assimilate spatio-temporal features and perform group activity recognition using three different input streams: RGB, optical flow, and 2D pose. 

In image classification, Dosovitskiy \etal ~\cite{dosovitskiy2020vit} showed that preliminary feature extraction using CNNs was not necessary. They proposed a pure transformer architecture called the Vision Transformer (ViT) that operated directly on sequences of image patches, or tokens, and found it performed very well on the image classification task. Arnab \etal ~\cite{Arnab_2021_ICCV} extended the Vision Transformer to video (ViViT) by extracting spatio-temporal tokens from input video. To handle the long sequences of tokens encountered in video, they further proposed factorising the input into spatial and temporal components to improve efficiency. ViVit achieved state-of-the-art accuracy on several action recognition benchmarks. 

In other areas of computer vision, Carion \etal ~\cite{detr} proposed the Detection Transformer (DETR) for object detection. Using an encoder-decoder Transformer to process CNN-extracted image features, they obtained comparable results to the popular Faster RCNN architecture~\cite{NIPS2015_14bfa6bb}. Li \etal ~\cite{Li_2021_CVPR} introduced two variants of an encoder-decoder Transformer architecture for single-stage (bottom-up) and two-stage (top-down) human pose estimation. In contrast to previous methods, their Transformer architectures regressed keypoints directly instead of using heatmaps.

\subsection{Computer vision based sports analytics}
Computer vision is currently being applied in many sports analytics problems. Problems such as sports event detection \cite{golfdb,Giancola_2018_CVPR_Workshops,Vats_2020_CVPR_Workshops}, player action recognition \cite{8781602,Cai_2019_CVPR_Workshops}, sports field registration \cite{namdar,sharma} and sports ball tracking \cite{Pidaparthy2019KeepYE,zhang_golf}  are being solved with the help of computer vision. McNally \textit{et al.} \cite{golfdb} use a hybrid CNN-LSTM network for golf swing sequencing and also introduce a new dataset for the same. Giancola \textit{et al.} \cite{Giancola_2018_CVPR_Workshops} introduce a new task of action spotting in soccer for finding anchors of game events in broadcast video. Pidaparthy \textit{et al.} \cite{Pidaparthy2019KeepYE} use  AlexNet \cite{NIPS2012_c399862d} to track the hockey puck in video by minimizing the mean-squared error (MSE) loss between the ground truth and predicted puck coordinates. Sharma \textit{et al.} \cite{sharma} perform field registration in soccer by computing the transformation between a broadcast image and static field model through nearest neighbour search. Cai \textit{et al.} \cite{Cai_2019_CVPR_Workshops} combine player stick and body pose with optical flow data to perform player level action recognition in ice hockey.

\subsection{Player identification from static images}
Before the advent of deep learning techniques, player identification from images was done with the help of hand crafted features. Although player appearance has been used to identify players in basketball \cite{Senocak_2018_CVPR_Workshops}, a player's jersey number remains a widely used feature for player identification due to its consistency in the game. Gerke \textit{et al.} \cite{gerke} was the first to use a CNN for identifying jersey numbers from player images. Vats \textit{et al.} \cite{vatsmmsports} introduce a multi-task loss function for identifying jersey numbers. Li \textit{et al.} \cite{li}  use a spatial transformer network to recognize jersey numbers from player images by warping the jersey number to suitable coordinates. Liu \textit{et al.} \cite{liu} augmented the Faster-RCNN \cite{NIPS2015_14bfa6bb} network with player pose information for detecting and recognizing jersey numbers from images. Gerke \textit{et al.} \cite{GERKE2017105} also merged their image-based jersey number identification system with player location features on the soccer field.

\subsection{Player identification from tracklets}
 Compared to inferring jersey numbers from static images, inferring jersey numbers from player tracklets has been found advantageous \cite{CHAN2021113891, vats2021player,jjl}. This is because the image sequences provide beneficial temporal information. Lu \textit{et al.} \cite{jjl} construct a conditional random field (CRF) consisting of feature nodes and identity nodes with appropriate connections and learn the CRF with weakly-supervised learning using a variant of expectation-maximization (EM). Chan \textit{et al.} use a network based on the LRCN network \cite{donahue}  to infer jersey numbers from player tracklets. The final tracklet scores are aggregated using a secondary CNN. Vats \textit{et al.} \cite{vats2021player} use 1D temporal convolutions to infer jersey numbers from player tracklets without the use of a secondary CNN. \par
 Our work is related to Lu \textit{et al.} \cite{jjl} as they also incorporate play-by-play as a prior during CRF training. We incorporate player shift information in a different way through multiplying the jersey number probability vector with binary shift vectors during inference (Section \ref{subsection:incorporating_player_shifts}). We test on a more diverse dataset consisting of 18 teams compared to two in Lu \textit{et al.} and  86 player identities compared to 24 (12 per team) in Lu \textit{et al.}.


\section{Methodology}
\subsection{Dataset}

The player identification tracklet dataset \cite{vats2021player} consists of $3510$ player tracklets. The dataset is obtained from 84 broadcast NHL videos. The tracklet bounding boxes and identities were annotated manually. The manually annotated tracklets simulate the output of a tracking algorithm. The average length of a player tracklet is $191$ frames. Note that the player jersey number is visible in only a subset of tracklet frames. The dataset is divided into 86 jersey number classes including one $null$ class representing no jersey number visible. The dataset is heavily imbalanced with the $null$ class consisting of $50.4\%$ of tracklet examples. \par
The training/testing split is done game-wise to avoid any in-game bias. $71$ videos are used for training/validation and $13$ videos are used for testing. The dataset contains  $2829$ training tracklets, $176$ validation tracklets and $505$ test  tracklets.


\subsection{Network architecture}
The input to the network is a temporal sequence of $m$ images    $\mathbf{T^m}= \{I_{i} \in \mathbb{R}^{3\times300\times300}\}_{i=1}^{m} $ sampled from a player tracklet   $\mathbf{T} = \{I_k: I_k \in \mathbb{R}^{300\times300\times3}\}_{k=1}^{n}$ of $n$ images. The $m$ images are randomly sampled from the tracklet $T$ serving as a form of data augmentation. The  sampling technique is discussed in Section \ref{section:sampling}.  The images $\mathbf{T^m}$ are passed through a 2D CNN (Resnet18 \cite{resnet}) to obtain $m$  features $\mathbf{F}=\{f_i \in \mathbb{R}^{512}\}_{i=1}^{m}$. The Resnet18 is pretrained on  static jersey number images using the image based jersey number dataset introduced by Vats \textit{et al.} \cite{vatsmmsports}.  The features $\mathbf{F}$ are input into a transformer encoder consisting of $l$ layers with $h$ multi-headed self-attention heads per layer. Each attention head has a constant dimension of $D_h \in  \mathbb{R}^{64}$. Positional encoding $p_i \in \mathbb{R}^{512}$ are added to the features $f_i$. Instead of using fixed positional encoding, the positional encoding is learned. As per the Vision transformer \cite{dosovitskiy2020vit}, a \texttt{[class]} token similar to BERT \cite{devlin-etal-2019-bert} is prepended to the CNN features $\mathbf{F}$. The state of the \texttt{[class]} token at the final transformer layer is fed to three multi-layer perceptron (MLP) heads consisting of a layernorm \cite{ba2016layer} and linear layer.  The output of the three MLP heads are three vectors. The first vector $p_{0} \in \mathbb{R}^{86}$ denotes the probability distribution of the predicted jersey number considering each jersey number in the dataset as a separate class. The other two vectors  $p_{1} \in \mathbb{R}^{11}$ and   $p_{2} \in \mathbb{R}^{11}$ denote the probability distribution of the first and second digit of the predicted jersey number. The one additional class in the 11-dimensional vectors $p_1$ and $p_2$ denotes the absence of a jersey number \par
We utilize the multi-task loss for jersey number recognition \cite{vatsmmsports} for training the network.  Concretely, we let $y_0 \in \mathbb{R}^{86}$ denote the ground truth vector for the holistic jersey number class, and we let $y_1 \in \mathbb{R}^{11}$ and  $y_2 \in \mathbb{R}^{11}$  denote the first digit and second digit ground truth vectors respectively. Let \begin{equation} \mathcal{L}_0 = -\sum_{i=1}^{86}y_0^i\log{p_0^i} \end{equation} be the holistic jersey number component of the  loss and  \begin{equation} \mathcal{L}_1 = -\sum_{j=1}^{11}y_2^j\log{p_1^j}  \end{equation} and  \begin{equation} \mathcal{L}_2 = -\sum_{j=1}^{11}y_1^j\log{p_2^j}  \end{equation} be the digit-wise losses.  Instead of using fixed weights for the three losses, the loss weights are learned using the technique introduced in Kendall \textit{et al.} \cite{Kendall_2018_CVPR}, with the overall loss $\mathcal{L}$  given by:

\begin{equation}
     \mathcal{L} =  \frac{1}{\sigma_1^2}  \mathcal{L}_{0} +  \frac{1}{\sigma_2^2}   \mathcal{L}_{1} +  \frac{1}{\sigma_3^2}   \mathcal{L}_{2} + \log(\sigma_1) + \log(\sigma_2) + \log(\sigma_3)
\end{equation}
where $\{ \sigma_i \}_{i=1}^{3}$ are trainable parameters. The overall network architecture is illustrated in Fig \ref{fig:overall_arch}.

\subsection{Training details}

For handling the severe class imbalance in the dataset, the $null$ class tracklets are sampled with a probability of $p_s = 0.1$ \cite{vats2021player}. The network is trained with Adam optimizer with an initial learning rate of $0.0001$ and a batch size of $16$. The learning rate is reduced by a factor of $\frac{1}{5}$ after $2500$ iterations and again after $5000$ iterations. Several data augmentation techniques such as random rotation by $\pm 10$ degrees, randomly cropping $300 \times 300$ pixel patches from the tracklet images and color jittering are used while training. Each augmentation technique is used on a per-tracklet basis instead of a per-frame basis. The experiments are preformed on two NVIDIA P-100 GPUs.

\subsection{Training through approximate labels}
\label{section:sampling}
\begin{figure}[t]
\begin{center}
\includegraphics[width=\linewidth, height = 4.4cm]{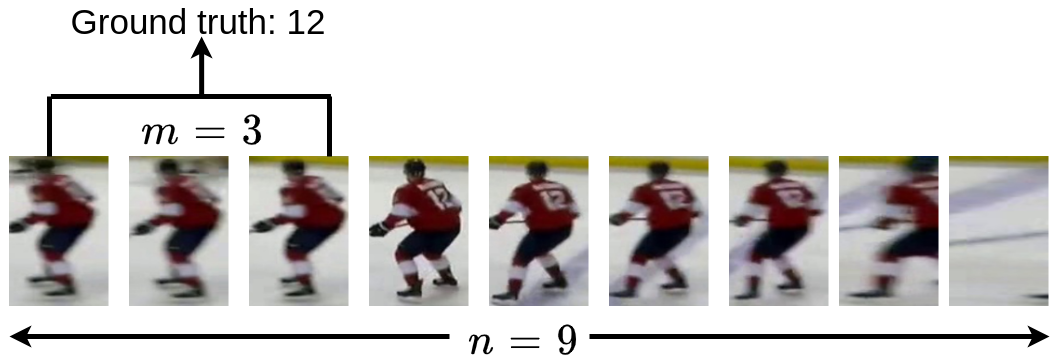}
\end{center}
  \caption{Toy exampling for a tracklet (of length $n=9$ frames) where sampling $m=3$ consecutive frames from the start leads to a sequence with ground truth $12$ with no jersey number visible. }
\label{fig:sampling_issue}
\end{figure}

The tracklets present in the training set can contain hundreds of frames such that the jersey number is only visible in a small subset of frames.  Previous approaches in the literature \cite{CHAN2021113891,vats2021player} sample a fixed number of frames randomly from a tracklet without any information of where the jersey number is actually visible. Therefore certain sampled tracklets with a non-null jersey number class may not have a jersey number visible. A toy example depicting such a scenario is shown in Fig. \ref{fig:sampling_issue}. This leads to inconsistent training signals which results in slow/unstable training as we demonstrate in experiments. To address this issue, we create frame-level labels indicating the frames in the tracklet where the jersey number is visible.  \par
To generate these frame level labels, let $\mathcal{M}$ be a model trained to predict a jersey number in static images and let $\mathbf{T} = \{I_k: I_k \in \mathbb{R}^{300\times300\times3}\}_{k=1}^{n}$ be a training tracklet consisting of $n$ images $I_k$. The model 
 $\mathcal{M}$ is run on every image $I_k$ to obtain the probability $p_k$ of whether a jersey number is visible in the image $I_k$. This gives $n$ probability scores $\{ p_k \in [0,1] \}_{k=1}^{n}$. The $n$ probability scores are thresholded with a binary threshold $\phi$ to obtain $n$ binary values  $\mathbf{B}= \{ b_k \in \{0,1\} \}_{k=1}^{n}$. The value of $b_k$ denotes the presence of jersey number in a tracklet frame.
 \begin{align}
     b_k &= 1 \; \text{if jersey number present in frame}\\
     b_k &= 0 \; \text{otherwise}
 \end{align}
The algorithm to obtain approximate labels in summarized in Algorithm \ref{algorithm:getB}.  The model $\mathcal{M}$ is a ResNet18 \cite{resnet} pretrained on a a jersey number dataset consisting of static images \cite{vatsmmsports}. \par
After precomputing $\mathbf{B}$, let $\mathbf{T^m} = \{ I_i \in \mathbb{R}^{300\times300\times3} \}_{i=l}^{l+m}$ where $l>=1$ and $l+m<= n$ be the $m$ images randomly sampled from a tracklet $\mathbf{T}$ for training.  The corresponding  $\mathbf{B^m}= \{ b_i \in \{0,1\}  \}_{i=l}^{l+m}$ where $l>1$ and $l+m<= n$ has at least one $b_i= 1$. This ensures that at least one image with a visible jersey number is present in the sampled tracklet. \par
For implementation, we let $\mathbf{I}$ denote the indices in the vector $B$ for which $b_k=1$. We randomly sample an index $start\_idx$ from $\mathbf{I}$ and then sample $m$ frames from the tracklet $\mathbf{T}$ starting from index $start\_idx$ to $start\_idx + m$.  A random offset $o \in [0,m)$ is subtracted from  $start\_idx$ to ensure that the sampled tracklet $\mathbf{T^m}$ may have a non-zero jersey number label at any sampled frame (and not necessarily always at the beginning). The algorithm is provided in Algorithm \ref{algorithm:sampleT_m}.



\begin{algorithm}[]
\setstretch{1}
\SetAlgoLined
\textbf{Input}: Player tracklet $\mathbf{T}$, Image-wise jersey number model $\mathcal{M}$ , Threshold $\phi$\\
\textbf{Output}: Frame for labels $\mathbf{B}$\\
\textbf{Initialize}: $\mathbf{B} = null$\\
\For{$I_k \quad \in \quad \mathbf{T} $}
{

   $p_k = \mathcal{M}(I_k)$
   
   \uIf{$p_k > \phi$}{
   
     $\mathbf{B}.\texttt{append}(1)$ 
    
    } 
   \Else{

    $\mathbf{B}.\texttt{append}(0)$ 
   }

}
\caption{Algorithm for creating approximate frame-wise jersey number labels.}
\label{algorithm:getB}
\end{algorithm}

\begin{algorithm}[]
\setstretch{1}
\SetAlgoLined
\textbf{Input}: Player tracklet $\mathbf{T}$, Frame-wise jersey number labels $\mathbf{B}$ , Sampling sequence length $m$\\
\textbf{Output}: Sampled tracklet images $\mathbf{T^m}$\\
\textbf{Initialize}: $\mathbf{T^m} = null$\\
\tcp{numpy function}
$\mathbf{I} = \texttt{np.where}(\mathbf{B} == 1)$

$start\_idx = \texttt{random\_sample}(\mathbf{I})$

$o = \texttt{randint}(m)$

$start\_idx = \texttt{max}(0, start\_idx - o)$

$T_m = \mathbf{T}[start\_idx: start\_idx+ m]$
\caption{Algorithm for sampling $m$ frames $\mathbf{T^m}$ for a tracklet $T$. }
\label{algorithm:sampleT_m}
\end{algorithm}

\subsection{ Incorporating player shifts}
\label{subsection:incorporating_player_shifts}
To incorporate player shifts for improving player identification performance, the game time in the video needs to be synced with the player shifts database, denoted by $\mathcal{S}$. $\mathcal{S}$ contains player shifts according to game time along with the corresponding jersey number and team affiliations.  To read game time from broadcast video clips, the EasyOCR\footnote{Found online at: \url{https://github.com/JaidedAI/EasyOCR}}  library was used.  Let $t_s$ denote the starting game time and $t_e$ denote the ending game time of a short video clip obtained using OCR. The player shifts $S^\prime$ that are present in the game time between $t_s$ and $t_e$ are extracted from the player shift database $\mathcal{S}$. The set $\mathcal{S}^\prime$ can be expressed as a union $S^\prime = S_h \cup S_a$ where $S_h$ and $S_a$ are the subsets of home and away shifts present in the  set $S^\prime$. Let the sets $\mathcal{H}$ and $\mathcal{A}$ denote the jersey numbers corresponding to  $S_h$ and $S_a$ respectively.   \par

Given a test video, player tracking and team identification are performed to obtain player tracklets \cite{vats2021player}. We assign a single jersey number probability vector $p_{jn}$ and team affiliation (home, away or referee) to each tracklet using the inference algorithm discussed in Vats \textit{et al.} \cite{vats2021player}. We then construct \textit{shift vectors} $v_h \in \mathbb{R}^{86}$ and $v_a \in \mathbb{R}^{86}$ that encode  the jersey numbers present in the home and away teams.  Let $null$ denote the no-jersey number class and $j$ denote the index associated with jersey number $n_j$ in  $p_{jn}$ vector. 

\begin{align}
    v_h[j] &=1, \text{if} \; n_j \in \mathcal{H}\cup\{null\} \\
    v_h[j] &= 0, \; otherwise 
\end{align}

\noindent similarly,  

\begin{align}
    v_a[j] &=1, \text{if} \; n_j \in \mathcal{A}\cup\{null\} \\
    v_a[j] &= 0, \; otherwise 
\end{align}
 
 Based on whether the player tracklet belongs to the home or the away team, the final player identity $Id$ is computed as 
 \begin{equation}Id=argmax(p_{jn} \odot v_h) \end{equation}
 
\noindent  (where $\odot$ denotes element-wise multiplication) if the tracklet belongs to the home team, otherwise, \begin{equation}Id=argmax(p_{jn} \odot v_a)\end{equation} if the player belongs to the away team.

 \begin{figure*}[t]
\begin{center}
\includegraphics[width = \linewidth]{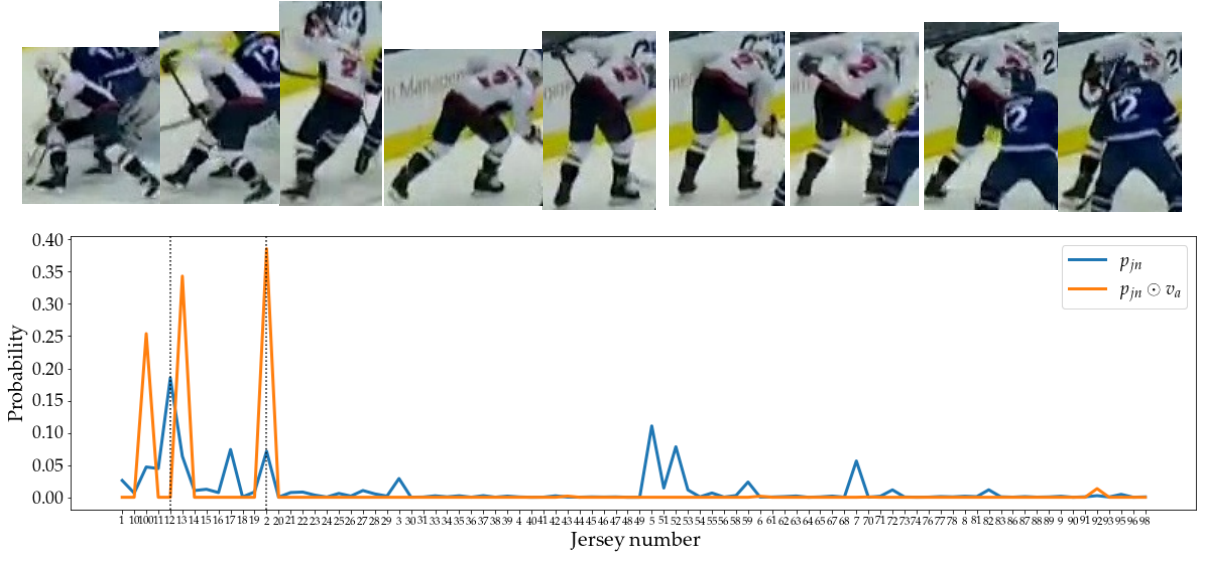}
\end{center}
  \caption{\textbf{Top row:} Example of a 'hard' tracklet where the ground truth jersey number $2$ is tilted. There is also a heavy occlusion with the opposition player with jersey number $12$. Note that the original tracklet contains 89 frames, however, only a subset of frames is shown here due to space constraints. \textbf{Bottom row:} For the tracklet shown in the top row, $p_{jn}$ is the probability of jersey number present in the tracklet (blue color). Orange color line is the normalized probability $p_{jn} \odot v_a$, i.e, the probability of jersey number multiplied by the shift vector $v_a$. For $p_{jn}$ the highest confidence value exists for jersey number $12$ (first vertical line from left), which is incorrect. Multiplying with the shift vector $v_a$ corrects the mistake by making the system focus only on the jersey number present in the away team during the game shift, after which the probability of the correct jersey number $2$ (second vertical line from left) becomes the greatest.}
\label{fig:shift_vec}
\end{figure*}

\section{Results}

\begin{table*}[!t]

    \centering
    \caption{Overall player identification accuracy for 13 test videos. The mean accuracy for identification increases by $5.95 \%$ after including the player shift data. }
    \footnotesize
    \setlength{\tabcolsep}{0.2cm}
    \begin{tabular}{c|c|c|c|c}\hline
  
       Video number & Ours w/ shift data & Ours w/ roster data & Ours w/o shift/roster data & Vats \textit{et al.} \cite{vats2021player} w/o shift/roster data \\\hline
     
     1  & $90.70 \%$ & $95.35\%$  &  $90.60\%$ & $90.60\%$ \\
      2  & $\textbf{91.43\%}$ &  $85.71\%$ &  $74.29\%$  & $57.1\%$ \\
       3  & $\textbf{87.72\%} $ &  $87.72\%$ &  $84.2\%$  & $84.2\%$ \\
        4  & $\textbf{80.00\%} $ & $76.0\%$  &  $72.00\%$  & $74.0\%$ \\
        5   & $\textbf{83.33\%} $ &  $83.33\%$ &  $81.48\%$  & $79.6\%$ \\
         6  & $ \textbf{90.00\%} $ & $90.0\%$  & $90.00\%$ & $88.0\%$ \\
          7  & $\textbf{85.07\%} $ & $80.60\%$  &  $73.13\%$  & $68.6\%$\\
      8  & $\textbf{93.75\%} $ & $93.75\%$   &  $91.6\%$ & $91.6\%$ \\
      
        9  & $\textbf{94.45\%} $ & $93.18\%$   & $88.6\%$  & $88.6\%$ \\
         10  & $\textbf{93.02\%}$ & $88.37\%$   &  $83.72\%$  & $86.04\%$\\
        11   & $\textbf{82.22\%}$ & $80.00\%$   &  $71.11\%$ & $44.44\%$ \\
        
        12  & $\textbf{84.85\%}$ & $84.85\%$   &  $84.85\%$ & $84.85\%$ \\
        13   & $\textbf{86.11\%}$ & $83.33\%$  &  $80.56\%$ & $75.0\%$ \\ \hline
        Mean & $\textbf{87.97\%}$  & $86.32 \%$ &  $82.02\%$  &  $77.9\%$
   
    \end{tabular}
    \label{table:pipeline_table}
\end{table*}

\label{section:result}
We compare the performance of the proposed network with Vats \textit{et al.} \cite{vats2021player}, which is the current state-of-the art on the dataset. The network performs better than Vats \textit{et al.}, demonstrating the effectiveness of the proposed approach.  The results are shown in Table \ref{table:result_comparison}. \par

We also re-implement Chan \textit{et al.} \cite{CHAN2021113891} from scratch due to unavailability of publicly-available code and dataset. The proposed approach obtains $10.1\%$ more accuracy than Chan \textit{et al.}. The reasons for better accuracy of the proposed approach compared to Chan \textit{et al.} are: (1) Chan \textit{et al.} use a temporal receptive field of only $16$ frames whereas the proposed approach has a more than double receptive field of $40$ frames. (2)  lack of data augmentation such as random rotation, color jittering in Chan \textit{et al.} (3) the dataset used in our work is half the size and much more skewed ($50.4\%$ $null$ class) compared to Chan et al. due to which their late fusion network overfits on our dataset. (4) Chan \textit{et al.} does not incorporate techniques to handle dataset class imbalance.  \par

\begin{figure}[t]
\begin{center}
\includegraphics[width=\linewidth, height = 5.2cm]{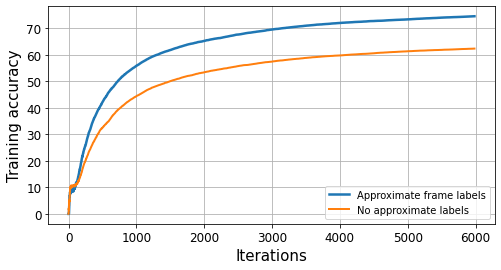}
\end{center}
  \caption{Training curves corresponding to a network with transformer layers $l=2$, attention heads per layers $h=8$ and training sequence length $m=40$ Training with approximate labels makes the network converge faster while training.}
\label{fig:training_curves}
\end{figure}

\begin{figure}[t]
\begin{center}
\includegraphics[width=\linewidth, height = 5.2cm]{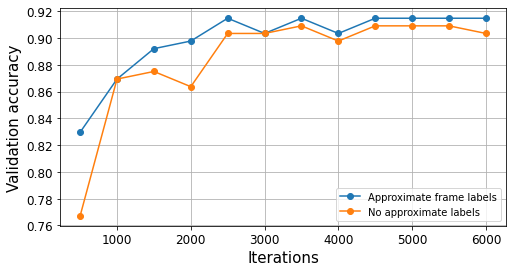}
\end{center}
  \caption{Validation accuracy curves corresponding to a network with transformer layers $l=2$, attention heads per layers $h=8$ and training sequence length $m=40$. The initial accuracy at iteration number $500$ is $6.2\%$ higher when training with approximate labels (blue color curve). The network also converges faster and obtains a higher accuracy value with approximate label based training.}
\label{fig:validation_curves}
\end{figure}

We also compare the proposed weakly-supervised training scheme making use of approximate labels to sampling frames randomly from any point in the tracklet (not using approximate frame labels) \cite{vats2021player, CHAN2021113891}.  The proposed scheme of training with the help of approximate labels improves the training convergence as illustrated in Fig. \ref{fig:training_curves}. The validation accuracy curves are shown in Fig. \ref{fig:validation_curves}. The reason for improved convergence with the proposed training scheme is that all the tracklet mini-batches sampled using approximate labels have the jersey number visible which results in a consistent training signal.

\subsection{Ablation studies}
\label{subsection:ablation_studies}
The number of transformer layers $l$, the number of attention heads $h$ and length of sequence for training/evaluation  $m$ are important parameters affecting the overall performance. Hence, an ablation study is performed to determine the best value for each parameter.

\subsubsection{Attention heads}
We perform an ablation study to determine to best value of the number of attention heads per transformer layer $h$. The values of $h \in  \{2,4,6,8,10\}$ were tested while keeping the number of transformer layers $l$ and sequence length for training/evaluation $m$ constant  $(l=2,m=30)$. The value of $h = 8$ showed the best performance with an accuracy of $83.6\%$ and a weighted F1 score of $84.2\%$. Table \ref{table:ablation_heads} shows the accuracy and F1 score values at the different values of $h$ tested. Using more than $8$ attention heads resulted in a performance decrease due to overfitting.

\begin{table}[!t]

    \centering
    \caption{Ablation study to determine the best value of attention heads per layer $h$ keeping number of layers $l$ and sequence length $m$ constant $(l=2,m=30)$. }
    \footnotesize
    \setlength{\tabcolsep}{0.2cm}
    \begin{tabular}{c|c|c}\hline
  
       $h$ & Accuracy &  F1 score\\\hline
     
     2  & $82.97 \%$ & $83.65 \%$ \\
      4  & $82.97 \%$ & $83.32 \%$   \\
       6  & $83.17\%$ & $83.74 \%$    \\
        8  &  $\textbf{83.37 \%}$ &  $\textbf{83.85\%} $   \\
        10   &  $82.38 \%$ & $82.90 \%$ \\

    \end{tabular}
    \label{table:ablation_heads}
\end{table}

\begin{table}[!t]

    \centering
    \caption{Ablation study to determine the best value layers $l$ keeping number of attention heads $h$ and sequence length $m$ constant $(h=8,m=30)$. }
    \footnotesize
    \setlength{\tabcolsep}{0.2cm}
    \begin{tabular}{c|c|c}\hline
  
       $l$ & Accuracy &  F1 score\\\hline
     
     2  &  $\textbf{83.37 \%}$ &  $\textbf{83.85\%} $   \\
      4  & $81.98\%$  & $82.74\%$ \\
       6  & $81.58 \%$ &    $82.17\%$ \\
        8  & $82.77 \%$ &  $83.17\%$   \\

    \end{tabular}
    \label{table:ablation_layers}
\end{table}

\begin{table}[!t]

    \centering
    \caption{Ablation study to determine the best value of training and evaluation sequence length $m$ keeping number of attention heads $h$ and number of layers $l$ constant $(h=8,l=2)$. }
    \footnotesize
    \setlength{\tabcolsep}{0.2cm}
    \begin{tabular}{c|c|c}\hline
  
       $m$ & Accuracy &  F1 score\\\hline
     
     10  & $81.58\%$  & $81.75\%$  \\
      20  & $83.37 \%$ & $83.76\%$  \\
       30  &  $83.37 \%$ &  $83.85\%$   \\
        40  & $\textbf{83.37} \%$ &  $ \textbf{84.14}\%$\\
        50   & $83.37\%$ & $84.07\%$  \\
    \end{tabular}
    \label{table:ablation_sequencelen}
\end{table}

\begin{table}[!t]

    \centering
    \caption{The result of the best performing model $(h=8,l=2,m=30)$ compared with the previous state-of-the-art on the dataset.}
    \footnotesize
    \setlength{\tabcolsep}{0.2cm}
    \begin{tabular}{c|c|c}\hline
  
       Model & Accuracy &  F1 score\\\hline
     
     Proposed  & $\textbf{83.37 \%}$ & $\textbf{84.14 \%}$ \\
      Vats \textit{et al.} \cite{vats2021player}  & $83.17 \%$ & $83.19 \%$   \\
    \end{tabular}
    \label{table:result_comparison}
\end{table}

\subsubsection{Transformer layers}

We  determine to best value of the number of transformer layers $l$ by testing $l\in\{2,4,6,8\}$ while keeping the number of attention heads per layer $h$ and the sequence length used for training/evaluation $m$ constant ($h=8,m=30$). From Table \ref{table:ablation_layers}, the best accuracy value of $83.37\%$ and F1 score of $83.85\%$ was obtained with $l=2$. The performance of the network declines after increasing the transformer layers from $l=2$ to $l=8$. This is because of overfitting since the number of parameters in the model increases around four times from $\sim3.2$ million when $l=2$ to $\sim12.6$ million when $l=8$ with no significant improvement in accuracy.

\subsubsection{Sequence length}
We determine the best value of the training and evaluation sequence length $m$ by keeping the transformer layers and number of attention heads per layer constant. The values of $m \in \{10,20,30,40,50 \}$. From Table \ref{table:ablation_sequencelen}, the lowest performance was shown by $m=10$ with an accuracy of $81.58\%$. Increasing $m$ to $20$ improved the accuracy and F1 score due to increase in receptive field of the network. However, the accuracy between $m=20$ to $m=50$ remained the same. The best performance was obtained by $m=40$ with an accuracy of $83.37 \%$ and F1 score of $84.14\%$.  Further increasing sequence length $m$ beyond $40$ did not improve performance.

\section{Result of incorporating player shifts}

We evaluate the network on the player tracklets obtained by running a tracking algorithm \cite{Braso_2020_CVPR,vats2021player} on the 13 test videos. This evaluation is different from the evaluation done in Section \ref{section:result} since the player tracklets are now obtained from the player tracking algorithm (rather than being manually annotated). The accuracy obtained by incorporating player shifts using OCR into player identification is compared to two baselines: (1) not incorporating any kind of roster/shift information, and (2) using player rosters available at the start of the game instead of player shifts \cite{vats2021player}. \par

From Table \ref{table:pipeline_table},  not using any shifts/roster data obtains a mean accuracy of $82.02\%$, that is $4.12\%$ greater than Vats \textit{et al.} \cite{vats2021player} . Incorporating player shifts obtains the best mean accuracy of $87.97\%$, which is $\sim6\%$ more than not using any shift or roster data. In fact, every video except the first video in the test set obtains equal or more accuracy when using the player shift data. This is because using player shifts helps the algorithm focus on a smaller subset of possible players present at a particular time. The lower accuracy of the first test video is due to inaccuracies in the shifts database.  Using the player roster obtains an accuracy $86.32\%$, which is just $1.65\%$ lower than the accuracy obtained when using player shifts, which demonstrates that even if player shifts are not available, using the available roster can provide performance comparable to using player shift data. Fig. \ref{fig:shift_vec} shows an example of a tracklet where incorporating player shifts corrects the prediction of the  model that does not use any shift or roster information.

\section{Conclusion}
In this paper, we introduced and implemented a transformer network for identifying players from player tracklets. We  introduce a novel, weakly-supervised training technique with the help of approximate labels to significantly speed up training. We also use a player shift database to significantly improve player identification accuracy on test videos. However, player identification is even more challenging when the jersey number of the player is not visible. Considering the fact that players in team sports such as ice hockey don't move randomly by follow roles such as defender, forward etc,future work will focus on improving player identification by incorporating a prior based on player positional data (e.g., left wing, center, right wing, defense, \etc ).

\section{Acknowledgment}

This work was supported by Stathletes through the Mitacs Accelerate Program and the Natural Sciences
and Engineering Research Council of Canada (NSERC). We also acknowledge Compute
Canada for hardware support.

{\small
\bibliographystyle{ieee_fullname}
\bibliography{egbib}
}

\end{document}